\documentclass[runningheads]{llncs}
\usepackage[T1]{fontenc}
\usepackage{graphicx}
\usepackage{float}
\usepackage{booktabs}
\usepackage{subcaption}

\begin{document}
\title{On Self-improving Token Embeddings}
%
%
%
\author{
Mario M. Kubek\inst{1}\orcidID{0000-0003-2641-2065} \and
Shiraj Pokharel\inst{1}\orcidID{0009-0003-5777-2749} \and 
Thomas Böhme\inst{2}\orcidID{0009-0005-3322-2456}\and
Emma L. McDaniel\inst{1}\orcidID{0000-0002-6179-5797} \and
Herwig Unger\inst{3}\orcidID{0000-0002-8818-3600} \and
Armin R. Mikler\inst{1}\orcidID{0000-0002-5253-144X}}

\institute{Georgia State University, Atlanta, GA, USA\\
\email{\{mkubek, emcdaniel10, amikler\}@gsu.edu} \email{; pokharel.shiraj@gmail.com}\and 
Technische Universität Ilmenau, Ilmenau, Germany\\
\email{thomas.boehme@tu-ilmenau.de} \and
FernUniversität in Hagen, Hagen, Germany\\
\email{herwig.unger@fernuni-hagen.de}}

\authorrunning{M. M. Kubek et al.}
%

%
\maketitle              
\begin{abstract}
This article introduces a novel and fast method for refining pre-trained static word or, more generally, token embeddings. By incorporating the embeddings of neighboring tokens in text corpora, it continuously updates the representation of each token, including those without pre-assigned embeddings. This approach effectively addresses the out-of-vocabulary problem, too. Operating independently of large language models and shallow neural networks, it enables versatile applications such as corpus exploration, conceptual search, and word sense disambiguation. The method is designed to enhance token representations within topically homogeneous corpora, where the vocabulary is restricted to a specific domain, resulting in more meaningful embeddings compared to general-purpose pre-trained vectors. As an example, the methodology is applied to explore storm events and their impacts on infrastructure and communities using narratives from a subset of the NOAA Storm Events database. The article also demonstrates how the approach improves the representation of storm-related terms over time, providing valuable insights into the evolving nature of disaster narratives.

\end{abstract}
\section{Motivation}
The assignment of meaningful representations to words and tokens is a crucial activity in many natural language processing (NLP) applications. Using the self-attention mechanism, large language models (LLMs) based on the Transformer architecture, such as BERT \cite{devlin-etal-2019-bert}, assign high-dimensional real-valued embedding vectors, usually referred to as word or token embeddings (hidden states), to input tokens. These hidden states are influenced and modified by the information of the surrounding tokens or text and represent an essential component of information processing in the Transformer network\setcounter{footnote}{0}\footnote{In technical contexts, the term `token' refers to individual units of text, such as words, punctuation marks, or subword units that are processed by a model. `Word', on the other hand, typically refers to a semantic unit used in natural language. In this article, the authors will use the terms `word embedding' and `token embedding' interchangeably, treating them as representations of these textual units regardless of their precise technical distinction.}. While they are propagated through the network, they are subject to constant changes in the multiple layers of the Transformer models, for example through the aforementioned self-attention mechanism and feed-forward networks. This approach is essential in order to capture subtle linguistic nuances, but it also means that the embeddings are highly contextualized due to these local influences and will result in very different hidden states even for thematically similar words (not just homonyms). 

In some applications, however, it is desirable to seek a more stable word and token representation, especially if the surrounding context is a thematically homogeneous document whose thematic orientation does not change. This could be the case in information retrieval systems that aim to semantically transform large collections of documents, in which the thematic orientation remains consistent across documents. Similarly, it can be relevant in document classification tasks in which documents belong to distinct categories, each with consistent thematic content.

To this end, it would make sense to calculate so-called static word embeddings using Word2Vec \cite{NIPS2013_9aa42b31}, FastText \cite{bojanowski-etal-2017-enriching}, or GloVe \cite{pennington2014glove} for the document set in question. While this approach is certainly feasible, its computational and time requirements vary depending on the size of the corpora. Furthermore, the generated word embeddings are uncontextualized, unlike those generated by an LLM. 

An alternative is the use of pre-trained word embeddings such as the vectors \cite{GoogleNewsVectors} calculated from the Google News Dataset. These embeddings cover three million words in the English language. However, the problem is that these embeddings stem from general-language sources, so that even and especially ambiguous terms are only assigned a single embedding that combines several contextual influences. This problem is exacerbated by the fact that word embeddings are often dominated by a word's most frequent meaning. The reason is that the frequency of word senses typically follows a power law \cite{kilgarriff2004dominant}, with one dominant sense and several less frequent, rarer meanings. However, within a specific domain, words typically appear in one single meaning. Therefore, many ambiguities can be avoided by selecting domain-specific texts.

Consequently, the out-of-the-box use of general language embeddings is not useful if terms in highly specialized or technical language documents are to be assigned a suitable representation. Even so, these general-language embedding vectors may provide a useful starting point for assigning stable and thematically oriented word embeddings to such terms in specialized documents and corpora.

In this article, the authors present a novel and fast approach to assign topic-specific word embeddings by iteratively refining pre-trained word embeddings using contextual information in topically oriented documents and corpora. The resulting word embeddings aim to better capture the document- and corpus-specific peculiarities and word usage patterns. In order to speed up their computation, this method does not rely on shallow or deep neural networks. This makes it possible to apply it in a real-time setting in which streams of textual data need to be processed as fast as possible. The authors regard this iterative refinement process as “self-improving,” since the embeddings evolve with each contextual update and increasingly reflect the semantics of the domain-specific corpus without requiring model retraining.

In the next section, the fundamentals of this method are introduced, with references to relevant literature. Section 3 then details its mathematical and technical aspects. Section 4 describes the experimental setup and presents the initial results. In Section 5, the method’s potential fields of application are discussed. Finally, Section 6 summarizes the article and explores the implications for future research and applications of the proposed method.

\section{Fundamentals}

The principle of distributional semantics, rooted in the theory of linguistic structuralism, suggests that words appearing in similar contexts have similar meanings. Embedding techniques like Word2Vec leverage this principle to create embeddings for words, tokens, and (sub)tokens \cite{biemann2022wissensrohstoff}. These high-dimensional real-valued vectors represent such linguistic units in a way that their geometric relationships in the vector space capture -- among other aspects -- semantic similarities and differences.

In this space, words with similar meanings are located close to each other, reflecting the underlying distributional properties learned from large text corpora. Typically, a single embedding is assigned to a word, although it would be more sensible to assign polysemous and homonymous words different embeddings reflecting their diverse semantic orientations and word senses. 

Furthermore, static embeddings can only represent words that were present in the training corpus. Words that did not appear during training (out-of-vocabulary words) cannot be represented effectively per se. Thus, a model generated by Word2Vec, for example, is confined to the world of the given training data. This problem has been alleviated by using subword information in models like FastText \cite{bojanowski-etal-2017-enriching}, which breaks words into character n-grams, allowing the representation of unseen words based on their constituent parts. Another solution called \`{a} la carte embedding \cite{khodak-etal-2018-la} is able to induce embeddings for previously unseen words by relying on a linear transformation that is efficiently learnable using existing pre-trained word vectors and linear regression.

When working with thematically homogeneous texts and corpora, as mentioned in Section 1, the out-of-the-box use of pre-trained word embeddings is only of limited help because they do not adequately represent the meanings of (possibly technical) terms, which are influenced and shifted by their specific contexts and the topical orientation in the actual documents at hand. It is therefore of particular interest to technically determine the degree of these shifts in meaning (especially with respect to the terms' pre-trained general-language embeddings) and the extent to which they capture the document- or corpus-related characteristics of the content. 

In the literature, there are various useful solutions for constructing representations of larger textual structures, including methods like text-representing centroid terms (TRCs) \cite{kubek-unger-2016}, Doc2Vec \cite{mikolov-le-2014}, and sentence transformers such as SBERT \cite{reimers-gurevych-2019-sentence}, which aim to provide meaningful representations at this level.

In contrast, the approach discussed herein works at the term level and is aimed at supporting the exploration of content relationships in thematically homogeneous corpora.

\section{Computing Topic-specific Embeddings}

\subsection{Preliminary Considerations}

The method's general goal is to iteratively update embeddings of words and tokens based on their contextual usage. At the same time, the approach aims to capture the evolving meanings of words by keeping a history of the updated embeddings. The method relies on the existence of a set of general-language pre-trained word embeddings such as the vectors \cite{GoogleNewsVectors} calculated from the Google News Dataset. 

Since these pre-defined embedding vectors are contextually adjusted by the method, this can be considered a specific case of transfer learning. However, while the use of pre-trained embeddings is not strictly necessary, as the desired word embeddings could also be generated through randomized initialization and iterative updates, reliance on this resource is still preferred. This allows these embeddings to transfer their pre-trained syntactic and semantic knowledge from large amounts of readily available and unlabeled data to other representations, tailored to local contexts. As such, the pre-trained embeddings are used during the initial look-up of a word that appears in the input text. 

\subsection{The Working Principle}
The general working principle of the method is as follows: The words, or more precisely tokens, of a text or corpus $D$ are extracted as part of a tokenization process and filtered according to a predefined part of speech list. Optionally, stop words should be removed and lemmatization can be performed. 

The method employs a fixed general context window that remains constant throughout the update process. This window size $s$ is determined based on the specific application requirements. Each target token, whose updated embedding is sought, can appear in a set of context windows $C_t$. Although only one context window is moved over the list of input tokens, which could be restricted to the necessary criteria (e.g., parts of speech, lemmata), this window covers $s$ positions in the list at each time step. It is therefore justified to speak of different windows here. Generally, all tokens can become a target token.

Within a specific context window $c$, the most current word embeddings are extracted for each target token $t$ (in the middle of the current context window $c$, denoted as $e_{t_{current}}$), and its neighboring tokens $n$ in $c$, denoted as $e_{n_{current}}$. Then, for each context window $c$, the target token's embedding $e_t$ is updated using learning rate $\alpha$ (hyperparameter) as shown in formula 1.
\begin{equation}
e_{t_{new}} = e_{t_{current}} + \alpha \sum_{n \in c} e_{n_{current}}
\end{equation}

The resulting embedding vector should additionally be normalized, as shown in formula 2:
\begin{equation}
e_{t_{new}} = e_{t_{new}} / || e_{t_{new}} ||
\end{equation}

As mentioned before, not only the most recent embedding vector of target token $t$ should be stored, but its update history as well. This will enable a more comprehensive analysis of its topical development. 

In case the token's embedding is updated for the first time, its current embedding $e_{t_{current}}$ used will point to the pre-trained embedding vector in case it exists. If it does not, a zero vector will be used as a placeholder. In this case, the initial update will assign a token embedding that is induced by the (existing) neighboring token embeddings in the current context window. The method can therefore compensate for missing embeddings by data imputation, too. The strength of the embedding adjustment therefore depends in particular on the parameter $\alpha$ and the size of the context window.

\subsection{Conceptual Influences}

As previously mentioned, the embedding vector of a target word is updated with each occurrence by combining its current embedding vector with the embedding vectors of its neighboring words within the respective context. This approach was partly inspired by the graph-based method of evolving text centroids presented in \cite{kubek-unger-2018}, which generalizes and extends the concept of text-representing centroid terms (TRCs) \cite{kubek-unger-2016}. Centroid terms are single, meaningful words that semantically and topically characterize text documents, thereby serving as compact symbolic representations in automated text processing tasks. 

The study demonstrated that such characteristic terms can be identified through a continuous and incremental adjustment of their positions within a large, general-purpose knowledge graph, tailored to the specific circumstances of the analyzed text. This process naturally forms traces of centroids (centroid trails), representing the thematic evolution of these terms until their final centroid terms are established. In contrast, the method described herein leverages word embeddings and is designed to quickly and continuously update the representations of potentially all terms in the given texts. Even so, it is not intended to produce a comprehensive document representation, as it operates solely at the term level. However, it retains the update history of a word's representation, thereby enabling an exploratory analysis of a term's topical shifts, similar to what can be achieved with centroid traces.

This approach also shares similarities with the self-attention mechanism found in Transformer-based neural network architectures to some extent. The similarity arises from the way both methods dynamically adjust representations based on the surrounding context. In the self-attention mechanism, the model computes attention scores that weigh the importance of each word relative to others in the sequence, allowing it to capture contextual relationships effectively. Similarly, in the method discussed here, the embedding of a target word is updated by incorporating information from its neighboring words, ensuring that its representation is continuously adapted to reflect the context it appears in. This iterative, context-driven adjustment of representations helps capture nuanced semantic shifts in a way that is conceptually akin to how self-attention works in Transformers. Both techniques aim to enhance the understanding of word meanings by accounting for their immediate linguistic environment, with the key difference residing in the underlying operations used to achieve this context-dependent adaptation.

\section{Experimental Results}

To demonstrate the usefulness of the method presented, a series of experimental results will be presented in this section. First, it will be shown that the method using pre-trained general language word embeddings can be applied to compute semantically meaningful adapted word embeddings for terms in thematically homogeneous texts and corpora, which are significantly more expressive than the original embeddings. Additionally, it is demonstrated that the method consistently generates embeddings that align closely with the topical area of the corpus. 

All of the following results are based on the ``GoogleNews-vectors-negative300” general-language dataset \cite{GoogleNewsVectors}, which contains 300-dimensional, pre-trained embedding vectors derived from the Google News dataset using Word2Vec \cite{NIPS2013_9aa42b31}. The plain text of the documents was extracted and tokenized, whereby only the parts of speech adjectives, adverbs, nouns, proper nouns, and verbs were allowed for the analysis (general-language stop words were removed, too). Also, all tokens, except proper nouns, were converted to lowercase, and lemmatization was performed. In case of multi-word expressions, only the individual components were extracted, not the multi-word expressions (connected with e.g. an underscore) as a whole.

\subsection{A First Example}

To illustrate the method's working principle qualitatively, the result of the analysis of the Wikipedia article ``abuse case” \cite{wikipedia-abuse-case} in the topical area of secure software engineering and threat modeling will first be interpreted. To parameterize the method, the size $s$ of the context window was set to 19 (the article's mean sentence length is 19.76) and the learning rate $\alpha$ to 0.01. The method was executed twice in succession (2 epochs).

\begin{table}[H]
  \centering
  \caption{Most similar terms to `abuse' and their similarity scores}
  \label{tab:abuse}
  \setlength{\tabcolsep}{12pt} 
  \begin{tabular}{cc|cc}
    \toprule
    Presented method & Score & Orig. pre-trained vectors & Score\\
    \midrule
     case & 0.98 & sexual\_abuse & 0.76\\
     use & 0.96 & Abuse & 0.67\\
     diagram & 0.94 & abusers & 0.63\\
     UML & 0.88 & abuses & 0.63\\
     security & 0.87 & abused & 0.63\\
     requirement & 0.86 & abuser & 0.62\\
     system & 0.86 & mistreatment & 0.61\\
     misuse & 0.85 & abusing & 0.61\\
     behaviour & 0.82 & maltreatment & 0.61\\
     term & 0.73 & sexual\_misconduct & 0.57\\
  \bottomrule
\end{tabular}
\end{table}

\begin{table}[H]
  \centering
  \caption{Cosine similarity values between updated embedding vectors for the terms most similar to `abuse' and their respective original pre-trained vectors after the latest update}
  \label{tab:cosine_similarity_latest_update}
  \setlength{\tabcolsep}{12pt} 
  \begin{tabular}{cc}
    \toprule
    Updated Term Vector & Cosine Similarity\\
    \midrule
     case & 0.33\\
     use & 0.68\\
     diagram & 0.61\\
     UML & 0.13\\
     security & 0.63\\
     requirement & 0.75\\
     system & 0.69\\
     misuse & 0.78\\
     behaviour & N/A\\
     term & 0.82\\
  \bottomrule
\end{tabular}
\end{table}

The resulting Table~\ref{tab:abuse} distinctly illustrates the divergence in the lists of terms considered most similar to the query term `abuse' based on the presented method's modified word vectors and the original pre-trained vectors. Similarity scores, truncated to two decimal places, were determined using cosine similarity. The method's results are influenced by the specific context of the analyzed dataset, leading to terms that reflect that context. Conversely, the original pre-trained vectors did not yield comparable results, which suggests a focus on terms related to more general associations with `abuse'. In addition, many of these terms are syntactically similar to the query term. These desired, context-specific results demonstrate how the method can yield relevant semantic relationships tailored to the specific dataset. 

Furthermore, Table~\ref{tab:cosine_similarity_latest_update} shows cosine similarity values between updated embedding vectors for the terms most similar to `abuse' and their original pre-trained vectors. Low cosine similarity values for terms such as `case' (0.33) and `UML' (0.13) indicate significant changes in their embeddings, which reflect considerable shifts from the original vectors. Notably, the term `behaviour' (British spelling variant) is marked as N/A because it was not found in the pre-trained model. For the term `abuse' (not included in the table) itself, the same calculation was performed. Here, the cosine similarity value obtained was 0.40, which also indicates a significant shift in its term vector.

\subsection{Analyzing Storm Events in the US}

The NOAA Storm Events Database \cite{noaa-storm-event-database} is a comprehensive collection of severe weather events documented by the National Oceanic and Atmospheric Administration (NOAA) in the United States. It provides detailed information on various types of extreme weather occurrences, such as tornadoes, hurricanes, floods, thunderstorms, and winter storms. Of particular interest are the often included event narratives, which offer detailed descriptions of the weather phenomena, impacts, and local conditions associated with a particular event. These narratives are particularly well-suited for methods that aim to extract key terms and uncover significant patterns in textual disaster records, as demonstrated by recent work on unsupervised key term extraction \cite{McDaniel-2023}.
Due to their topical directedness, these event narratives are an interesting subject of analysis for the presented method, too. Therefore, as an example, combined narratives of storm events from 1993 were analyzed. 

After pre-processing, the token list to analyze contained 3,466 unique tokens (out of the original 141,206) elements from 4,418 documents (out of the original 8,664) in the dataset that actually contain storm event narratives. Notably, only 2,872 of those unique tokens have a representation in the set of the original pre-trained vectors used. This clearly shows that the mentioned out-of-vocabulary problem is indeed to be addressed. The method's parameters were set as follows: the size $s$ of the context window was set to 13 (aligned with the corpus' mean sentence length of 14.72), and the learning rate $\alpha$ was set to 0.01, 0.075, and 0.15 respectively. The method was executed twice in succession (2 epochs).

\begin{figure}[H]
  \centering
  \includegraphics[width=0.90\linewidth]{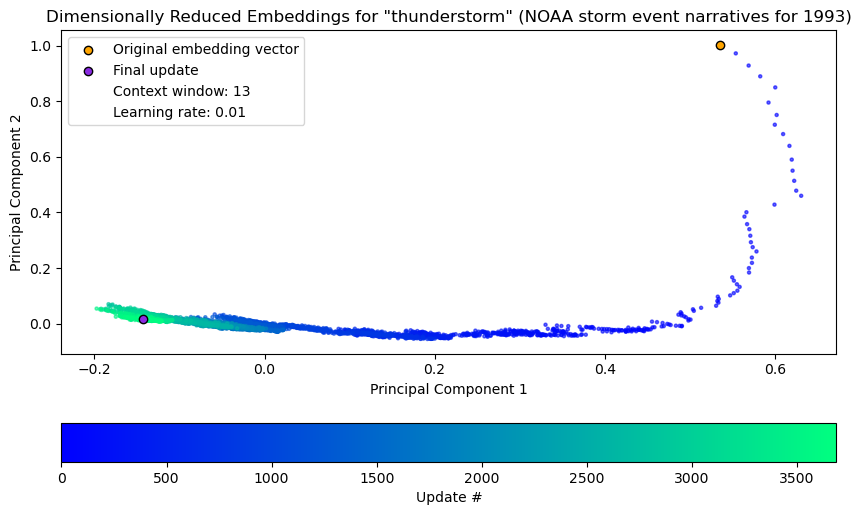}
  \caption{Two-dimensional representation of the evolving embedding vector for the term `thunderstorm' with learning rate 0.01 and context window size of 13.}
  \label{fig:thunderstorm1}
\end{figure}

\begin{figure}[H]
  \centering
  \includegraphics[width=0.90\linewidth]{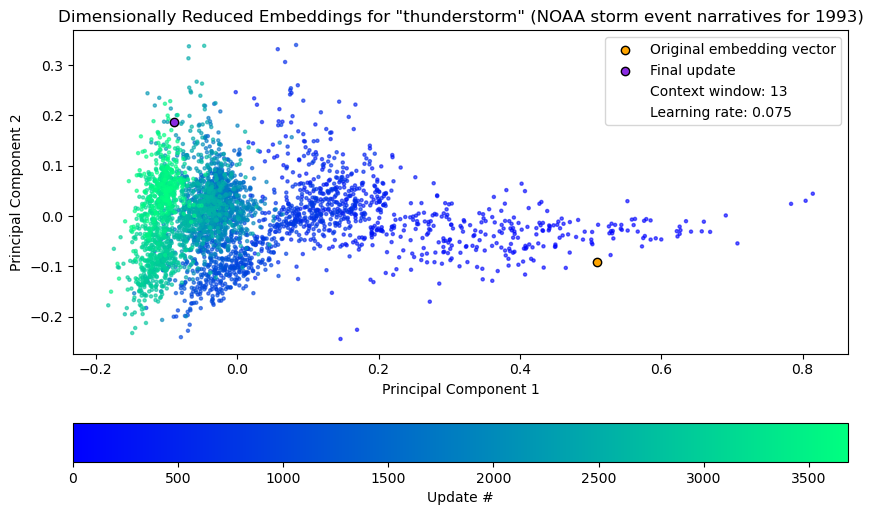}
  \caption{Two-dimensional representation of the evolving embedding vector for the term `thunderstorm' with learning rate 0.075 and context window size of 13.}
      \label{fig:thunderstorm2}
\end{figure}

\begin{figure}[H]
  \centering
  \includegraphics[width=0.90\linewidth]{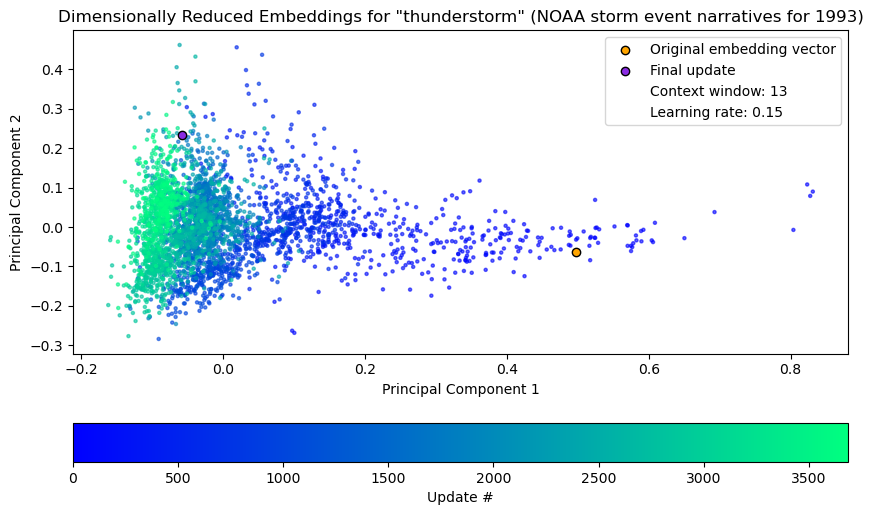}
  \caption{Two-dimensional representation of the evolving embedding vector for the term `thunderstorm' with learning rate 0.15 and context window size of 13.}
  \label{fig:thunderstorm3}
\end{figure}

Figures~\ref{fig:thunderstorm1},~\ref{fig:thunderstorm2}, and~\ref{fig:thunderstorm3} illustrate the transformation of the embedding vector for the term `thunderstorm' into a two-dimensional space using Principal Component Analysis (PCA). 
Principal Component 1 (on the x-axis) captures the primary direction of variation in the token embeddings, representing differences between words associated with natural phenomena, such as thunderstorms, earthquakes, or floods. As a result, the embedding for `thunderstorm' will be positioned along the Principal Component 1 axis, close to other climate-related terms. Principal Component 2 (on the y-axis) represents a more specific direction of variation in the embeddings. This component captures finer distinctions, such as the intensity of the terms. For instance, `thunderstorm' would be placed closer to terms like `tornado' or `avalanche' and farther from milder weather terms like `breeze.' 

Additionally, the figures demonstrate how the embedding vector evolves relative to its original representation from the pre-trained dataset as more text is processed. Specifically, 3,686 embedding vectors were generated from 1,843 occurrences of the term `thunderstorm' extracted from the corpus. The influence of different learning rates on the stability of these embeddings is clearly observable, with varying rates affecting how the vector shifts over time.

A learning rate $\alpha$ of 0.075 and 0.15 results in significant fluctuations influenced by the local context of the term. This observation aligns with the behavior of neural networks during gradient descent, where large learning rates lead to instability. Conversely, a much smaller learning rate $\alpha$ of 0.01 leads to embedding vectors with less fluctuation, a behavior consistently observed across various experiments and datasets. Therefore, a smaller learning rate is recommended, as it leads to a more stable representation of the term in the embedding space. It is worth noting that the method's efficiency, attributed to the application of linear algebra operations, ensures rapid corpus processing. On an i7-12650H processor, the analysis of the 1993 storm event narratives corpus will take approximately four seconds. 

To qualitatively assess the results of the presented method, Table~\ref{tab:thunderstorm} lists the five terms most similar to the query terms `thunderstorm', `tornado', and `emergency' for both the presented method and the originally pre-trained vectors. It can be observed that the word embeddings returned by the method are more meaningful, specific, and representative of the given topically homogeneous corpus. For example, the most similar embedding vectors represent terms that point to the impacts and accompanying characteristics of storm events and related terms. 

\begin{table}[H] 
  \caption{Most similar terms to `thunderstorm', `tornado', and `emergency' according to the embeddings created by the presented method (top) and the original pre-trained vectors (bottom)}  
  \label{tab:thunderstorm}
  \centering
  \setlength{\tabcolsep}{12pt} 
  \begin{tabular}{ccc}
    \toprule
    \multicolumn{3}{c}{Presented method} 
     \\
    thunderstorm & tornado & emergency \\
    \midrule
     storm & crop & management \\
     cluster & public & office \\
     mph & baseball & barn \\
     spotter & school & department \\
     limb & thunderstorm & roof \\
     quarter & lightning & county \\
     time & windshield & afternoon \\
     fence & farmstead & community \\
     outbuilding & ground & Lubbock \\
     sign & police & Skywarn \\
  \bottomrule
\end{tabular}
\\ 
\vspace{\baselineskip}
  \begin{tabular}{ccc}
    \toprule
    \multicolumn{3}{c}{Orig. pre-trained vectors} \\
    thunderstorm & tornado & emergency \\
    \midrule
     thunderstorms & twister & Emergency\\
     severe\_thunderstorm & tornadoes & emergencies\\
     severe\_thunderstorms & tornados & emer\_gency\\
     rainstorm & twisters & emegency\\
     thundershower & F3\_tornado & emergeny\\
     thunder\_storms & EF3\_tornado & nonemergency\\
     heavy\_rain & funnel\_cloud & ambulance\\
     downpour & F2\_tornado & nonemergencies\\
     rain & tornado\_touched & technician\_Tim\_Rujan\\
     Thunderstorms & EF5\_tornado & EMERGENCY\\
  \bottomrule
\end{tabular}
\end{table}  

In contrast, the terms most similar to the original vectors are more general and primarily indicate syntactically similar expressions which do not support deeper topic analysis. Hence, even though the original vectors are employed, for the purpose of corpus exploration, the method's returned and modified word embeddings are better suited as they properly capture the underlying documents' topical directions. Furthermore, while the method shares some resemblance with the graph-based and symbolic method of evolving centroid terms \cite{kubek-unger-2018}, the resulting embedding vectors can more accurately capture the syntactic and semantic nuances of the terms. In the analyzed corpus, the average cosine similarity of 0.56 for the mentioned 2,872 unique tokens' embeddings to their original embeddings suggests that the method generates embeddings that greatly differ from the original ones.

\begin{figure}[htbp]
    \centering
    \begin{subfigure}{0.45\textwidth}
        \centering
        \includegraphics[width=\linewidth]{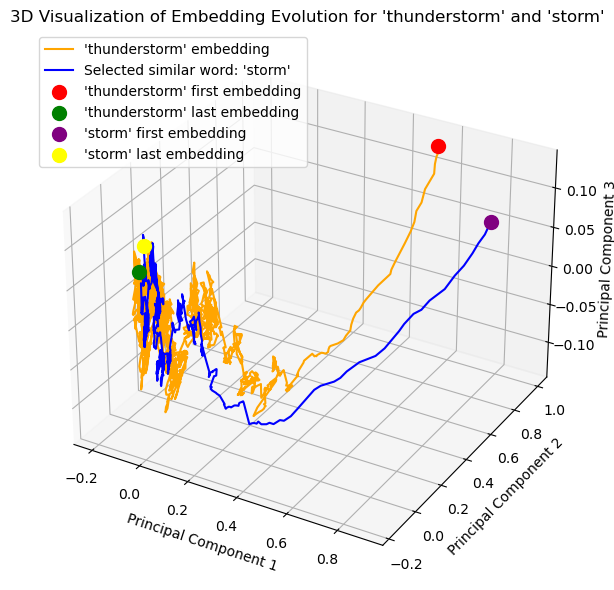}
        \caption{thunderstorm vs. storm}
        \label{fig:subfig1}
    \end{subfigure}
    \hfill
    \begin{subfigure}{0.45\textwidth}
        \centering
        \includegraphics[width=\linewidth]{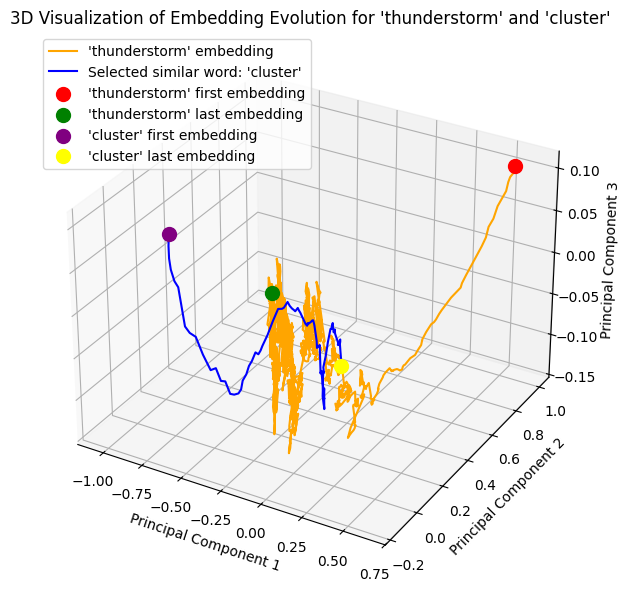}
        \caption{thunderstorm vs. cluster}
        \label{fig:subfig2}
    \end{subfigure}
    
    \vspace{1em} 

    \begin{subfigure}{0.45\textwidth}
        \centering
        \includegraphics[width=\linewidth]{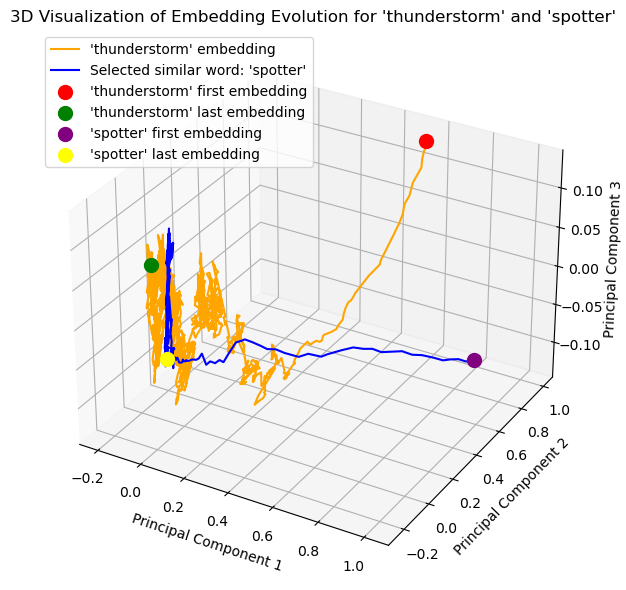}
        \caption{thunderstorm vs. spotter}
        \label{fig:subfig3}
    \end{subfigure}
    \hfill
    \begin{subfigure}{0.45\textwidth}
        \centering
        \includegraphics[width=\linewidth]{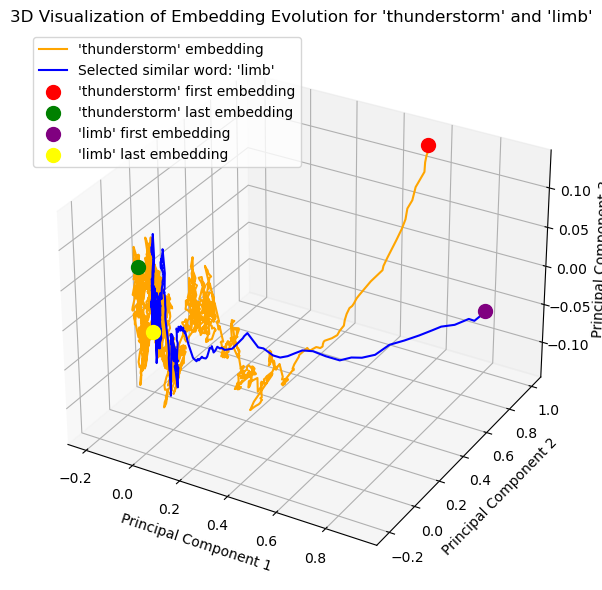}
        \caption{thunderstorm vs. limb}
        \label{fig:subfig4}
    \end{subfigure}
    
    \vspace{1em} 

    \begin{subfigure}{0.45\textwidth}
        \centering
        \includegraphics[width=\linewidth]{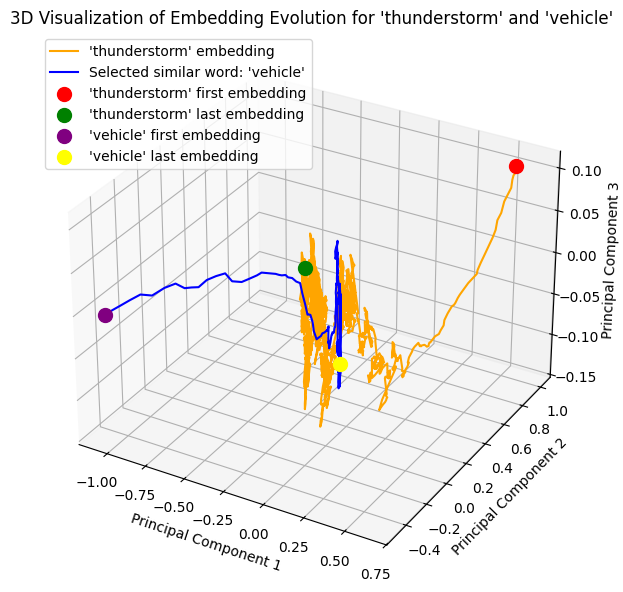}
        \caption{thunderstorm vs. vehicle}
        \label{fig:subfig5}
    \end{subfigure}
    \hfill
    \begin{subfigure}{0.45\textwidth}
        \centering
        \includegraphics[width=\linewidth]{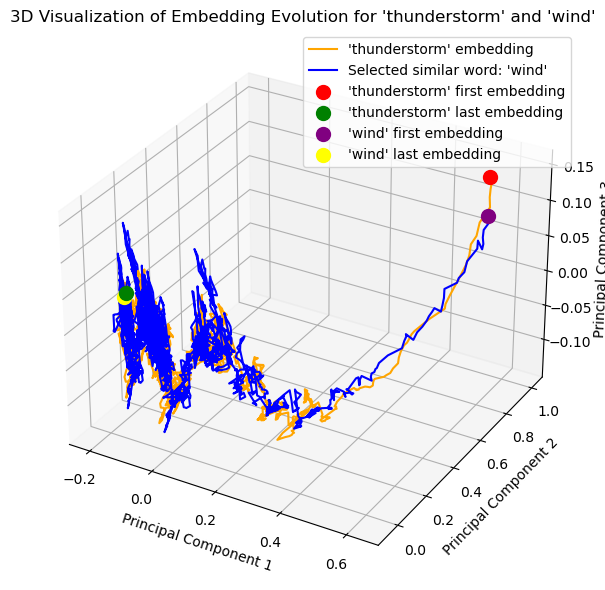}
        \caption{thunderstorm vs. wind}
        \label{fig:subfig6}
    \end{subfigure}

    \caption{3D Visualization of the temporal evolution of word embeddings for selected term pairs using `thunderstorm' as a reference}
    \label{fig:mainfigure}
\end{figure}

Figure~\ref{fig:mainfigure} presents six 3D PCA projections of the temporal evolution of embeddings for term pairs, each representing `thunderstorm' and a manually selected comparison term. The key parameters of the method were set to a context window size of $s=13$ and a learning rate of $\alpha=0.01$. The first four selected terms -- `storm,' `cluster,' `spotter,' and `limb' -- are among the most similar to `thunderstorm' after applying the method. The embeddings for `vehicle' and `wind', as represented in subplots~\ref{fig:subfig5} and~\ref{fig:subfig6}, exhibit distinct characteristics, which are explained below.

These embeddings evolve over time as they are updated based on storm-related event descriptions extracted from the NOAA Storm Events Dataset \cite{noaa-storm-event-database}. The principal components in the plots capture the most significant variance in the embedding space after dimensionality reduction. Each subplot uses color-coded trajectories and markers to indicate:
\begin{itemize}
    \item The trajectory of `thunderstorm' (orange) and that of a given selected word (blue).
    \item The initial embedding position (red for `thunderstorm,' purple for the selected word).
    \item The final embedding position after all updates (green for `thunderstorm,' yellow for the selected word).
\end{itemize}
  
Subplot~\ref{fig:subfig1} illustrates that the embedding for `storm' remains closely aligned with `thunderstorm' throughout the updates, with both trajectories evolving in parallel. The initial positions (red for `thunderstorm,' purple for `storm') are already close, and the final positions (green for `thunderstorm,' yellow for `storm') remain in proximity, which indicates strong semantic stability. The closely aligned trajectory suggests minimal divergence and reinforces that `storm' consistently co-evolves with `thunderstorm' in the dataset. It also suggests that these terms are used interchangeably in the given storm event descriptions.

Subplot~\ref{fig:subfig2} shows that the embedding for `cluster' starts farther from `thunderstorm' than `storm' but gradually shifts closer over time. While the initial positions (red for `thunderstorm,' purple for `cluster') indicate a considerable semantic distance, the final positions (green for `thunderstorm,' yellow for `cluster') show increased proximity. This change suggests that `cluster' becomes more contextually associated with `thunderstorm' as updates are applied, likely due to its frequent co-occurrence in storm-related reports where storm clusters are described.

Subplots~\ref{fig:subfig3} and~\ref{fig:subfig4} present the temporal evolution of the embeddings for `spotter' and `limb', each in relation to `thunderstorm'. Both exhibit similar trajectories. The embeddings start farther from `thunderstorm' compared to more directly related meteorological terms like `storm', and their movements indicate a weaker but still notable semantic association. While both words shift closer to `thunderstorm' over time, their final positions remain relatively distant, which indicates a more indirect contextual relationship. `Spotter' is linked to storm observation and reporting, while `limb' may appear in contexts describing storm-related impacts such as damage or injuries. The observed embedding shifts reflect how these terms interact with storm descriptions in the dataset without being core meteorological concepts.

Subplot~\ref{fig:subfig5} shows that the embedding for `vehicle' initially lies in a semantically distant region from `thunderstorm', which reflects their distinct meanings. Over time, however, their embeddings shift closer together, suggesting an increasing contextual association. This shift is likely driven by their frequent co-occurrence in descriptions of storm impacts, particularly damage to vehicles caused by thunderstorms. While `vehicle' does not inherently belong to meteorological terminology, its embedding adapts to reflect its relevance in the given corpus, where storm-related damages and disruptions play a significant role.

In subplot~\ref{fig:subfig6}, the embedding for `wind' initially occupies a position very close to `thunderstorm', and their trajectories remain nearly identical throughout the updates. This strong alignment indicates a stable semantic relationship, likely due to the intrinsic connection between wind and thunderstorms in meteorological contexts. Unlike other words, which undergo noticeable shifts, `wind' and `thunderstorm' evolve in parallel. This pattern indicates that their contextual association is well-established in the dataset and remains consistent over time.

A key observation is that while some words, such as `wind', remain semantically close to `thunderstorm' from the beginning and exhibit nearly identical trajectories, others, like `cluster' and `vehicle', start from more distant regions and gradually shift closer due to their contextual co-occurrence with storm-related events. This effect is particularly pronounced in homogeneous corpora, such as the analyzed storm-related reports, where words related to storm impacts (e.g., `vehicle,' `limb') or observation (`spotter') undergo embedding adjustments due to their frequent co-occurrence with `thunderstorm' in descriptive contexts.

These results reinforce the idea that word embeddings capture not only static semantic relationships but also context-dependent shifts driven by corpus updates. This study provides insights into how specialized vocabulary in domain-specific datasets stabilizes over time and reflects both inherent semantic relationships and contextual shifts.

\section{Applications}

The following applications are hypothetical examples from research and practice. They illustrate how the proposed method aligns with various requirements and objectives in different contexts. Future work will investigate these solutions in detail to evaluate their effectiveness and optimize their implementation.

\subsection{Corpus Exploration: Term Clustering and Topic Tracking}

The method of self-improving embeddings holds significant benefits for corpus exploration, particularly in tasks involving term clustering and the temporal tracking of topics or terms. By maintaining a history of embedding updates, this approach enables researchers and practitioners to observe how the semantic representations of terms evolve across a corpus or over time. This, in turn, allows for the derivation and investigation of new research questions and hypotheses.

\subsubsection{Term Clustering}
Through iterative refinement, the embeddings generated by this method naturally group semantically or topically related terms in the vector space. This facilitates the identification of term clusters that reveal underlying themes or concepts within a corpus. For example, in a dataset of research articles, terms related to a specific scientific field can converge into cohesive clusters, enabling an automated exploration of thematic patterns.

\subsubsection{Topic and Term Evolution}
The ability to retain and analyze embedding history adds a unique dimension to topic tracking. By examining changes in the embeddings of specific terms, it becomes possible to monitor shifts in their contextual usage. This is particularly valuable in longitudinal studies, where the emergence or evolution of topics can be traced across different time periods. For instance, in social media analysis, this technique can help capture how the usage of trending terms adapts to evolving public discourse.

In both applications, the context-dependent adjustment of embeddings ensures that the resulting representations are finely tuned to the corpus-specific semantics. This adaptability, combined with the method’s computational efficiency, makes it a suitable tool for exploring large, diachronic and topically homogeneous text corpora.

\subsection{An Interactive Q\&A Chatbot Built on a Local Knowledge Base}

Providing effective employee support through a generative AI-powered chatbot requires more than just retrieving documents based on keywords; it necessitates a nuanced understanding of the organization's internal knowledge base. This is especially challenging when the knowledge base consists of diverse file formats like Word, Excel, PDF, and PowerPoint, each tailored to specific organizational purposes. General-purpose embeddings, utilized via an enterprise-licensed or open-source vector database, do not adequately capture the domain-specific nuances embedded within such documents. These static embeddings are trained on fixed datasets, making them ill-suited to adapt to evolving organizational knowledge and contexts. Consequently, responses based on these embeddings risk being generic or outdated, which can diminish user engagement.

The presented method addresses these limitations by continuously refining embeddings based on the content and context of the internal knowledge base. This ensures that the chatbot can dynamically adapt to changes in organizational knowledge, enabling it to offer responses that are precise, context-aware, and tailored to employee queries.

\subsubsection{Conceptual Search Through Semantic Transformation}

By mapping a document’s terms and the document itself into a high-dimensional semantic vector space, a semantic transformation is performed. This allows the chatbot to conduct conceptual searches by gaining a deeper understanding of the textual content. In this space, relationships between words, concepts, and themes are encoded, enabling the chatbot to identify and retrieve relevant information.

Although a semantic transformation does not necessarily require symbolic methods like TF-IDF (term frequency–inverse document frequency) \cite{salton-yan-1973}, incorporating term weights can improve the process by highlighting domain-specific terms while downweighting common, less-informative ones (e.g., `the' or `system'). This hybrid approach merges the symbolic precision of TF-IDF with the rich contextual understanding of word embeddings, empowering the chatbot to uncover both explicit and implicit connections within the knowledge base. Key benefits of this approach include:

\begin{itemize}
    \item \textbf{Context-Aware Retrieval:} Instead of merely identifying documents containing the queried keywords, the chatbot retrieves documents that are conceptually aligned with the query. By leveraging refined embeddings and semantic mapping, the chatbot ensures its responses reflect the broader meaning and relevance of the query.
    \\
    \item \textbf{Conceptual Queries with Weighted Embeddings:} Using weighted combinations of word embeddings, powered by techniques like TF-IDF, the chatbot can process complex queries involving multiple terms or themes. For example, a search for `increasing revenue streams' might retrieve documents discussing profitability, growth strategies, or competitive analysis, even when the exact term `revenue' is absent.
\end{itemize}

Through conceptual search, the chatbot not only finds documents that match a query explicitly but also identifies those that share a deeper semantic alignment. This makes it a powerful tool for exploring and navigating the organization’s internal knowledge base with greater precision and insight. Beyond internal use cases, such capabilities are highly relevant for community services as well, where public-facing chatbots or local information platforms can benefit from context-sensitive, continuously adapting embeddings -- e.g., to support emergency response, citizen inquiries, or evolving municipal knowledge bases that require up-to-date information representation.

\subsubsection{Adapting to Evolving Knowledge}

By analyzing query logs and iteratively updating embeddings, the chatbot ensures that its semantic space evolves alongside the knowledge base. This dynamic adaptation captures emerging trends and updates domain-specific embeddings, keeping the chatbot relevant and effective over time. With this approach, the generative AI chatbot transitions from a static query-response system to an intelligent assistant capable of conceptual understanding, empowering employees with precise and insightful answers.




\subsection{Continuous Deployment and Integration of Data Pipelines}
As new word embeddings are generated, they can be directly integrated into existing data pipelines without downtime. A versioning system can ensure that the data pipeline can check-point to previously used embeddings if new updates do not work as expected. A/B testing can be performed to compare the performance of the old and new embeddings on a subset of organizational use cases involving queries, ensuring that updates improve the data pipeline’s overall performance without introducing performance degradation.

Once the updated embeddings are deployed, real-time monitoring can assess the pipeline’s improved performance. Metrics such as response accuracy, user satisfaction ratings, and query resolution time should be tracked.

\section{Conclusion}

In this article, the authors presented a novel and fast method for deriving topic-specific word embeddings. It captures the terms' topical directions in topically homogeneous corpora by considering their individual contexts. Designed to continuously update these word representations, it also enables the analysis of topic shifts. In the context of the showcased storm event analyses, it was demonstrated that this feature supports the dynamic assessment of the impacts and risks of these events for specific locations, providing a comprehensive understanding of their evolving effects. Furthermore, the applicability of the proposed method to evolving language data and task-specific requirements was highlighted. For instance, the authors showed that this approach allows embeddings to incrementally improve, resulting in more robust NLP systems capable of handling complex linguistic and contextual variations. This makes the method particularly valuable for real-time, community-centered applications such as local information services, emergency coordination, or public knowledge platforms. Future research work will investigate how the method can be combined with symbolic knowledge representations or lightweight neural models to support structured reasoning and domain adaptation, and to enable integration into large-scale, multilingual, and dynamic information systems that rely on interactive and agent-based components as core elements.

\section{Acknowledgments}

The authors sincerely thank Mr. Chandra Kiran Guntupalli, Master's student in Computer Science at Georgia State University, Atlanta, GA for his invaluable efforts in the preparation and interpretation of the 3D visualizations, which have greatly contributed to this work.

%
%
%
%

\end{document}